# A Conjoint Application of Data Mining Techniques for Analysis of Global Terrorist Attacks

## Prevention and Prediction for Combating Terrorism

Vivek Kumar[1] [Orcid], Manuel Mazzara[2], Maj. Gen. (Rtd.) Angelo Messina[2], JooYoung Lee[2]

[1] National University of Science and Technology-MiSiS, Moscow, Russian Federation, vivekumar0416@gmail.com

[2] Innopolis University, Kazan, Russian Federation, m.mazzara@innopolis.ru a.messina@innopolis.ru,j.lee@innopolis.ru

**Abstract.** Terrorism has become one of the most tedious problems to deal with and a prominent threat to mankind. To enhance counter-terrorism, several research works are developing efficient and precise systems, data mining is not an exception. Immense data is floating in our lives, though the scarce availability of authentic terrorist attack data in the public domain makes it complicated to fight terrorism. This manuscript focuses on data mining classification techniques and discusses the role of United Nations in counter-terrorism. It analyzes the performance of classifiers such as Lazy Tree, Multilayer Perceptron, Multiclass and Naïve Bayes classifiers for observing the trends for terrorist attacks around the world. The database for experiment purpose is created from different public and open access sources for years 1970-2015 comprising of 156,772 reported attacks causing massive losses of lives and property. This work enumerates the losses occurred, trends in attack frequency and places more prone to it, by considering the attack responsibilities taken as evaluation class.

## 1 Introduction

"Our responsibility is to unite to build a world of peace and security, dignity and opportunity for all people, everywhere, so we can deprive the violent extremists of the fuel they need to spread their hateful ideologies."

– António Guterres, Secretary-General of the United Nations.

Uncountable losses of lives and assets happen every year all around the globe. The interesting thing to note is sometimes some organizations claims their hand behind or involvement in its execution while sometimes nobody takes the responsibilities of the attack. The objective of this work is getting more concrete idea, by analyzing all the incidents to build an intelligent model which can be engaged institutive predictions. It has been observed that terrorists usually do not claim responsibility for attacks and from the statistics says that terrorist groups claim credit for only one out of seven attacks [1-5].

Before we go deeper from physiology point of view, the questions that can come across the mind are: - What is the need for claiming an attack? What difference claiming and attack or not will make? The possible answer is claiming an attack publicly in the present time is a way to attract attention, to be taken seriously, to terrorize a whole range of people and different countries at the same time. Aftermath of Paris and Brussels attacks shocked not only France and Belgium, but Europe, America, and the whole world. Similarly, there are several reasons for not claiming an attack, such as to avoid retaliation or to jeopardize the hideout bases. Also, it is evident that they do not need the publicity in the regions of their stronghold. With the increasing reach to cutting edge-technology, and evolution of new ways of advance warfare, the whole planet is endangered. The motivations of terrorist organizations are clear, the question that naturally arises is: what can be done to strengthen counter terrorism? The answer is predicting the attacks beforehand and prevent them in time. However, the path to execution is not as simple as the answer seems. It requires several functional divisions like homeland security, intelligence agency, armed forces working together to connects the dots.

United Nation is also playing a vital role in combating terrorism. Peacekeeping operations have evolved to adapt and adjust to hostile environments, emergence of asymmetric threats and complex operational challenges that require a concerted multidimensional approach and credible response mechanisms to keep the peace process on track. The military component, as a main stay of a United Nations peacekeeping mission plays a vital and pivotal role in protecting, preserving and facilitating a safe, secure and stable environment for all other components and stakeholders to function effectively. The Global Counter-Terrorism Strategy in the form of a resolution and an annexed Plan of Action (A/RES/60/288) that included an overview of the evolving terrorism landscape, recommendations to address challenges and threats, and a compilation of measures taken by Member States and United Nations entities to fight against terrorism. And composed of 4 pillars:

- Addressing the conditions conducive to the spread of terrorism.
- Measures to prevent and combat terrorism.
- Measures to build states' capacity to prevent and combat terrorism and to strengthen the role of the United Nations system in that regard;
- Measures to ensure respect for human rights for all and the rule of law as the fundamental basis for the fight against terrorism.

In this, ongoing system artificial intelligence, machine learning and other software technology play a key role. Though fundamental remains the same- information. Correct information or accuracy and precision in the information is no doubt the game changer. As data science is now into everything for instance healthcare, finance, business and what now, its possible crucial contribution cannot be denied. This work contributes for the same to connect data mining and counters terrorism [6-8]. Data mining, text mining, sentiment analysis, machine learning systems, and predictive analyt-ics are techniques which can be utilized to recognize and battle terrorism. Text mining helps us to decipher unknown information by extracting it automatically from different written sources [9-15].

## 2 Classification Techniques Used for Data Mining

Classification is a form of data analysis that extracts models describing important data classes. For example, we can build a classification model to categorize if the tumor is benign or malignant for patients going under treatment of possible breast cancer. The process is shown for the possible suspect data in a Figure 1 and 2 in simplified way. Such analysis helps to provide us with a better understanding of the data at large scale. It is used for several research applications including fraud detection, terrorism prediction, target marketing, performance prediction, medical diagnosis, finance, weather prediction, business intelligence, homeland security. Several approaches are used for classification of datasets, as there are numerous techniques for classification and rule extraction. Classification algorithms can be categorized as probabilistic or non-probabilistic classifiers, Binary, and Multiclass classifier. Data classification is a two-step process namely learning step (where a classification model is constructed) and second step called classification step (where the model is deployed to predict the class labels of given data).

A sample, X is represented by an n-dimensional attribute vector, X = $(x_1,x_2, , x_n)$ depicting $n$ measurements made on the sample from $n$ attributes of database, ($A_1$, $A_2$,...... , $A_n$). Each attribute represents a feature of X. The accuracy of a classifier for a given set of test samples is the percentage of test set samples that are correctly classified by the classifier. The associated class label of each test sample is compared with the learned classifier's class prediction for that sample [16-24]. A brief working insight of few prominent classifiers are presented below.

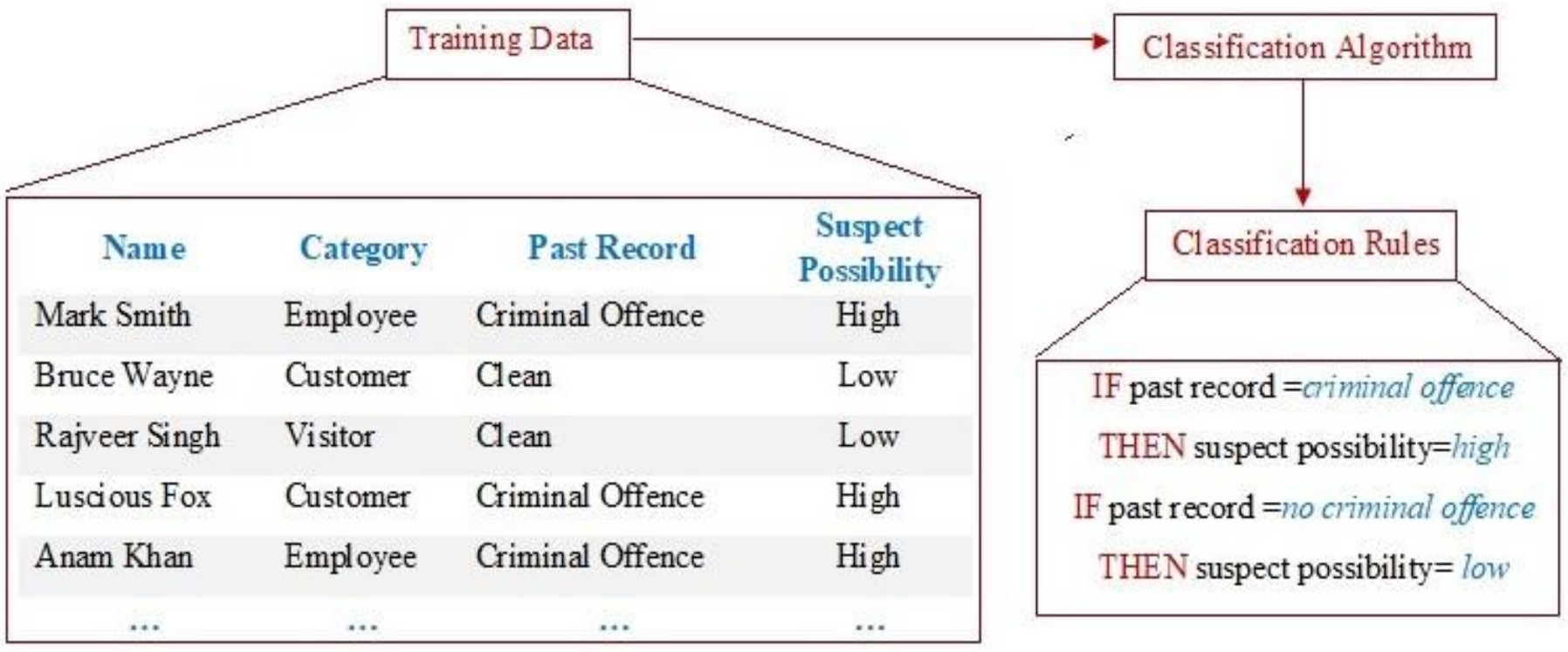

**Fig. 1.** Learning: Training data are analyzed by a classification algorithm and here the class label attribute is suspect possibility, and the classifier is represented in the form of classification rules.

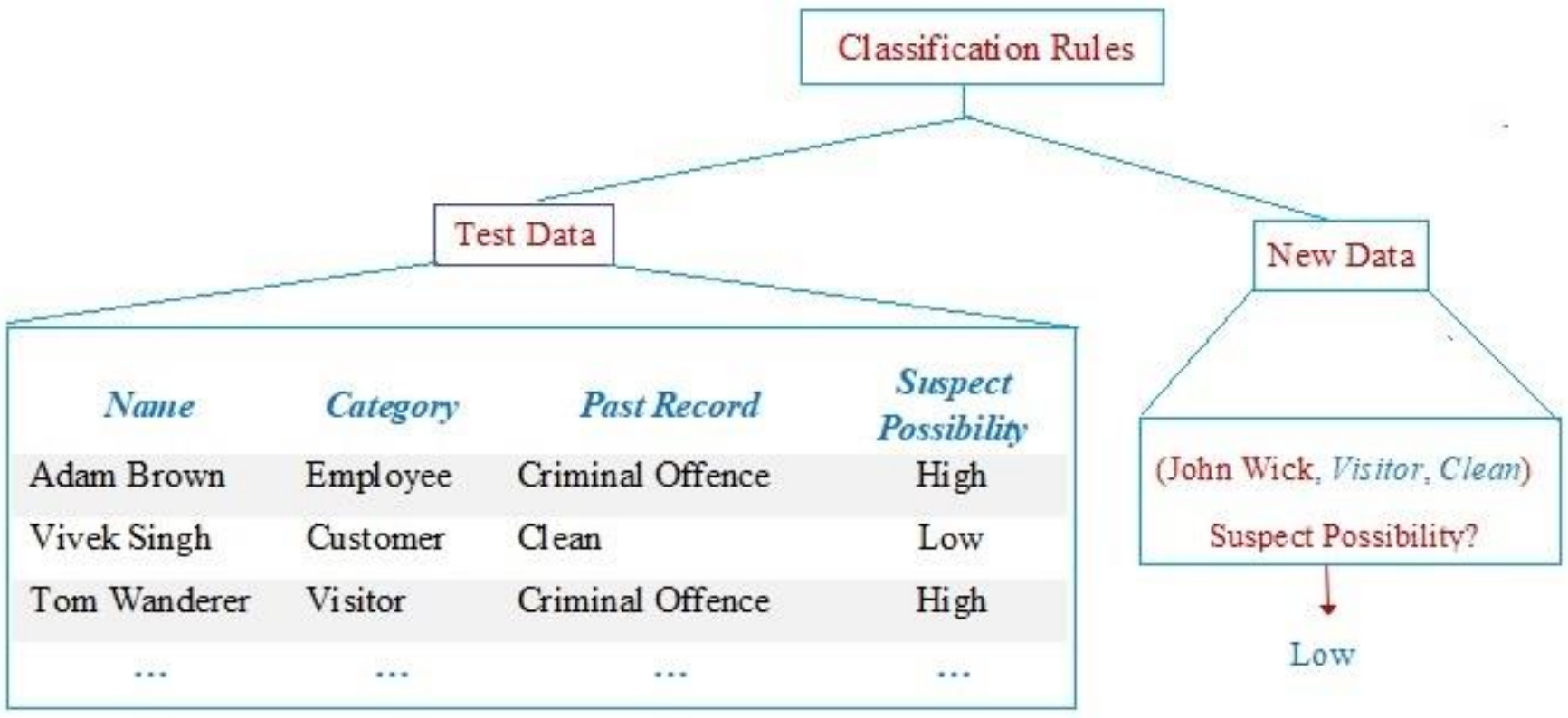

**Fig. 2.** Classification: Test data are used to estimate the accuracy of the classification rules. If the accuracy is considered acceptable, the rules can be applied to the classification of new data samples.

### 2.1 Naïve Bayes

It is a classification technique based on Bayes' Theorem, in which it assumes that the presence of a particular feature in a class is unrelated to the presence of any other feature. This model is easy to build and particularly useful for very large data sets. Along with simplicity, Naive Bayes is known to outperform even highly sophisticated classification methods. It is easy to understand the algorithm in some steps. First, it converts the data set into a frequency table. After this create likelihood table by finding the probabilities. Lastly using the equation of Naïve Bayes, to calculate the posterior probability for each class. The outcome of prediction is the class with greatest posterior probability. In a probabilistic model, using Naïve Bayes theorem the conditional probability can be given by:

$$(C_K|x) = \frac{P(x|C_K)P(C_K)}{P(x)} \tag{1}$$

In other words, it can be written as Posterior = (prior × likelihood) | evidence. Under the conditional distribution over the class variable C, the equation can be given by:

$$P(C_K|x_1, \dots., x_n) = \frac{1}{Z}\left(P(C_K)\prod_{i=1}^{n} P(|x_i|C_K)\right) \tag{2}$$

where the evidence (Z) = P(x) is scaling factor, only dependent upon $x_1, \dots, x_n$. Bayesian Classifier although have some complexities such as, it requires prior information of probabilities and in its absence, it is often predicted on the basis of background knowledge and earlier available data about original distributions.

## 2.2 Decision Tree

A decision tree is a flowchart-like tree structure, where each internal node (non-leaf node) denotes a test on an attribute, each branch represents an outcome of the test, and each leaf node (or terminal node) holds a class label. The topmost node in a tree is the root node. The construction of decision tree classifiers does not require any domain knowledge or parameter setting, and therefore is appropriate for exploratory knowledge discovery. Decision trees can handle multidimensional data. Their representation of acquired knowledge in tree form is intuitive and generally easy to assimilate by humans. The learning and classification steps of decision tree induction are simple and fast. In various computations, the characterization is executed recursively till every single leaf is immaculate, that is the data order which should be as flawless as would be prudent. The objective is a progressive observation of a decision tree until it gets adjusted of flexibility and precision. This method used the entropy that is the calculation of unstructured information. Here Entropy $\vec{X}$ is measured by:

$$\text{Entropy}\ (\vec{X}) = -\sum_{i=1}^{n} \frac{|Xi|}{|X|} log(\frac{|Xi|}{|X|}) \tag{3}$$

$$\text{Entropy}\ (i|\vec{X}) = \frac{|X|}{|X|} log(\frac{|Xi|}{|X|}) \tag{4}$$

Total gain can be represented by

$$\text{Total Gain} = \text{Entropy}\ (\vec{X}) - \text{Entropy}\ (i|\vec{X}) \tag{5}$$

## 2.3 Multilayer Perceptron

Multilayer Perceptron consists of a minimum of three layers of nodes and except the input nodes; each node is a neuron that uses a nonlinear activation function. It uses a supervised learning method known as back propagation for training. If a multilayer perceptron has a linear activation function in all neurons, then according to linear algebra any number of layers can be reduced to a two-layer input-output model. In perceptron learning happens by changing connection weights after each portion of data is processed, depending upon the amount of error in the output compared to the expected result. This is a supervised learning example, which is carried out by back propagation. The error in output node j in the nth data point is represented by

$$e_j(\text{n}) = d_j(\text{n}) - y_j(\text{n}) \tag{6}$$

where d is the target value and y is the value yielded by the perceptron. The node weights are managed on the basis of corrections that minimize the error in the entire output, given by

$$\text{E (n)} = \frac{1}{2}\sum_j e_j^2(n) \tag{7}$$

Using gradient descent, the change in each weight can be depicted as

$$\Delta\omega_{ji}(n) = -\eta\delta \frac{\delta E(n)}{\delta v_j(n)} y_i(n) \tag{8}$$

where $y_i$ is the output of the previous neuron and η is the learning rate, which is selected to ensure that the weights swiftly converge to a response, without oscillations. The derivative depends on the induced local field $v_j$, which itself varies. For an output node the, derivative can be simplified to

$$-\frac{\delta E(n)}{\delta v_j(n)} = e_j(\mathrm{n})\, \emptyset'(v_j(n)) \tag{9}$$

where $\emptyset'$ is the derivative of the activation function previously mentioned, which does not vary by itself. The analysis is more tedious for the change in weights to a hidden node, although it can be shown that the required derivative is

$$-\frac{\delta E(n)}{\delta v_j(n)} = \emptyset'(v_j(n)) \textstyle\sum_k \frac{\delta E(n)}{\delta v_k(n)} w_{kj}(n) \tag{10}$$

This depends on the change in weights of the nth nodes, which represent the output layer. In order to change the hidden layer weights, the output layer weights change according to the derivative of the activation function.

## 3 Methodology

The work itself starts from collection of data. All attempts were made to keep to data coherent and correct as much as possible. The second step is data preprocessing. It is a crucial undertaking and basic stride in text mining, and information retrieval. Data preprocessing is a data mining strategy that includes transforming raw data into a format suitable for experimental purposes. Data pre-processing stage is supposedly the most time-consuming of the whole knowledge discovery phase. The data is filtered after preprocessing by means of several filters such as linear and non-linear. All the attacks location with their corresponding latitude and longitude were embedded to the world map. The density plot snapshot is shown in Figure 3 and 4 all attacks in the Word, Middle East and North Africa (40422 attacks) respectively. This research work focuses on classification based upon class –Attack Responsibility for the dataset comprised of reported terrorist attacks from 1970-2015. The classification algorithms applied for knowledge discovery for this data are Decision Tree random forest, Lazy classifier IBK linear NN, Lazy classifier IBK Filtered Neighbor Search, Lazy tree, IBK, Ball Tree, Lazy classifier K-star, Multilayer Perceptron, Multiclass Classifier and Naïve Bayes. So the class we have chosen for analysis for terrorist's attacks and causalities is based upon the fact if an organization has taken credit for it or not or whether it been an anonymous attack.

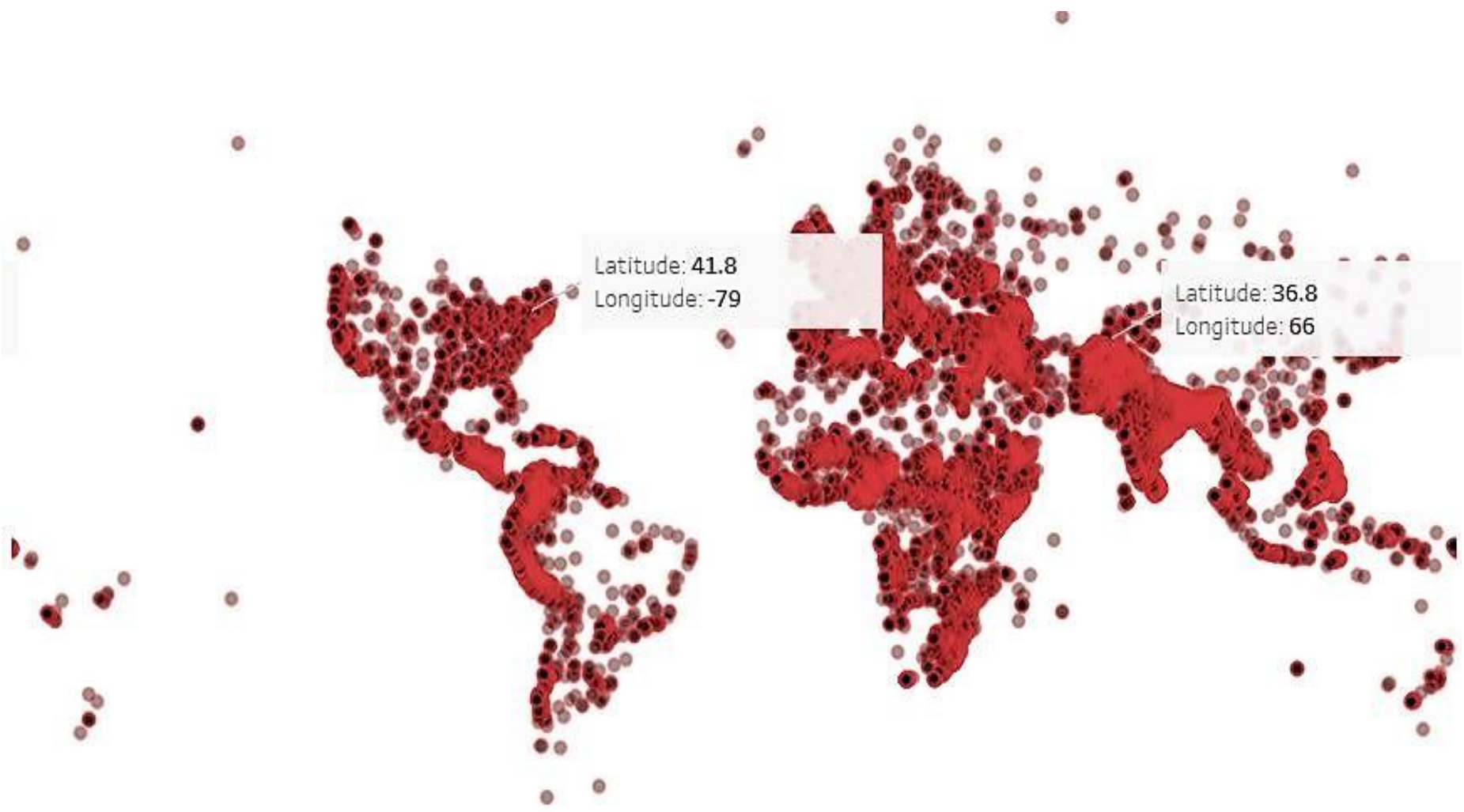

**Fig. 3.** Density distribution plot of all the terrorist attack (world) embedded with their latitude and longitude.

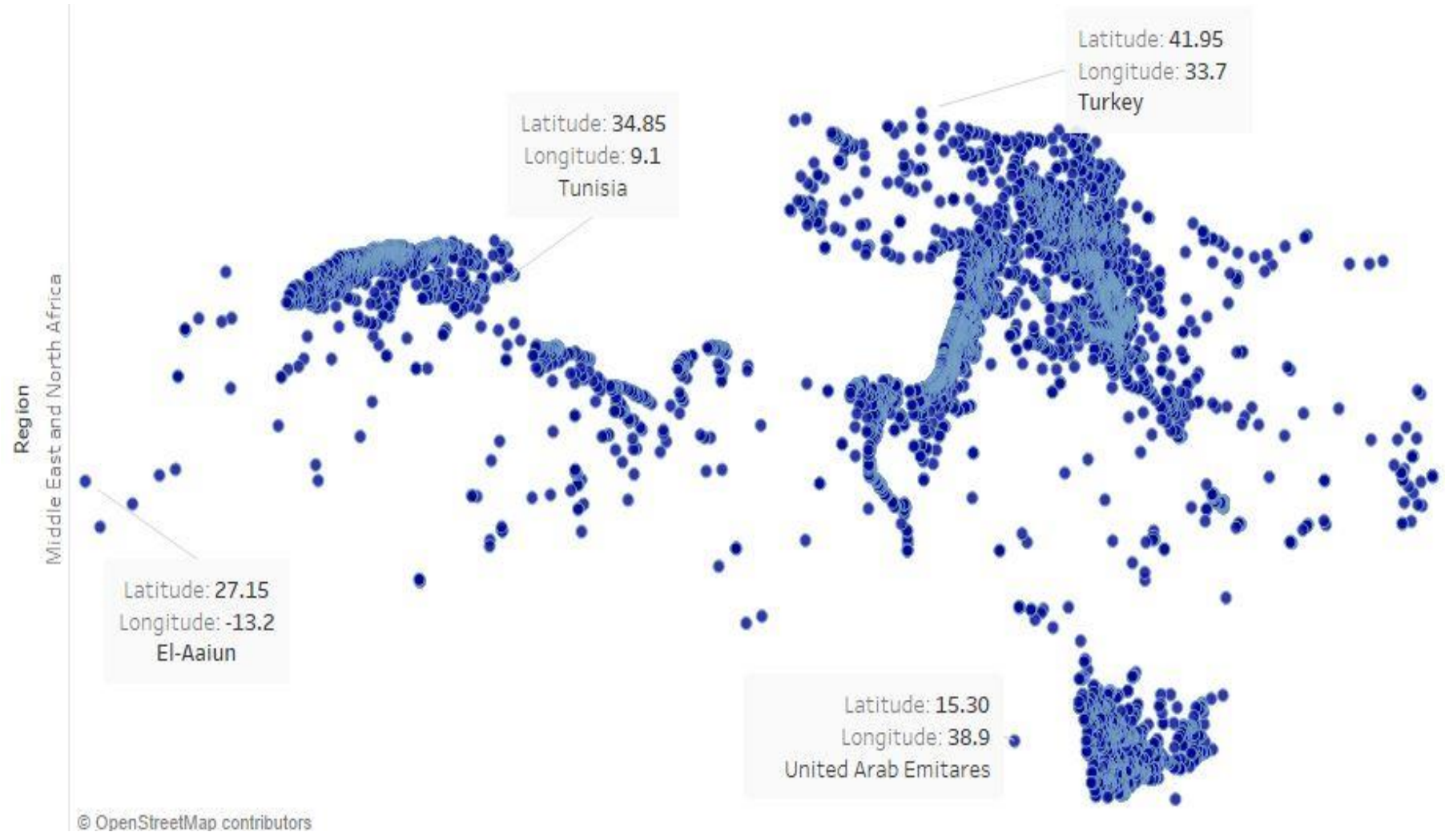

**Fig. 4.** Middle East & North Africa density distribution plot of 40422 terrorist attacks embedded with their latitude and longitude.

### 3.1 Description of Dataset

The terrorist attack data is obtained from the online data sources including sources released by the ministry of defense, government, private organizations and by collecting it individually. This data set is comprised of 156772 reported attacks all over the world, happened between the years 1970 to 2015. Analysis of the data after re-fining the raw data ends in nine attributes for categorizing data for this study. The attributes which were considered for this analysis were namely month of attack, year of the attack, Region, Weapon Type, Target, Attack Type, Data Source and Property Loss.

**Table 1.** Numerical distribution of experimental dataset.

| | | | |
|---|---|---|---|
| Attack Type | AT-1 | Armed Assault | 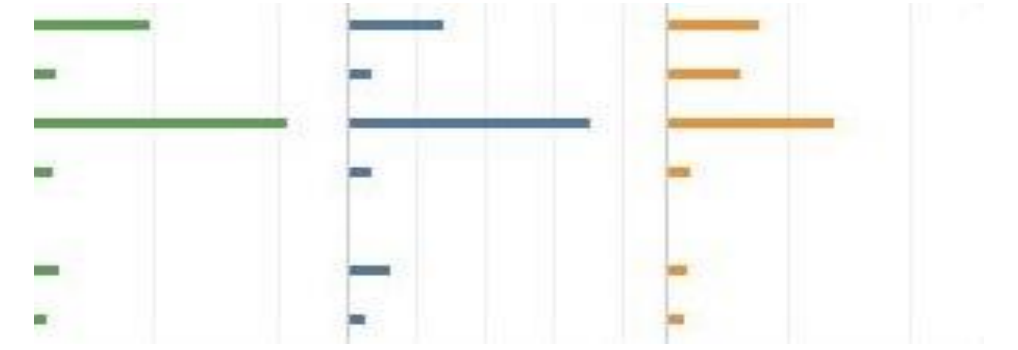 |
| | AT-2 | Assassination | |
| | AT-3 | Bombing | |
| | AT-4 | Facility/Infrastructure | |
| | AT-5 | Hostage(Kidnapping) | |
| | AT-6 | Hijacking | |
| | AT-7 | Others | |
| | | | |
| Property Loss | S | Major | 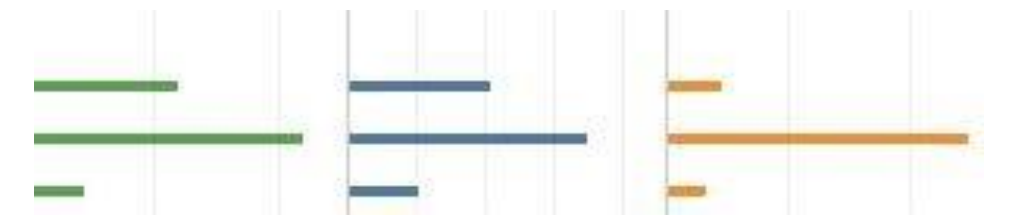 |
| | M | Moderate | |
| | L | Minor | |
| | U | Unknown | |
| | | | |
| Region | R1 | Central America & Caribbean | 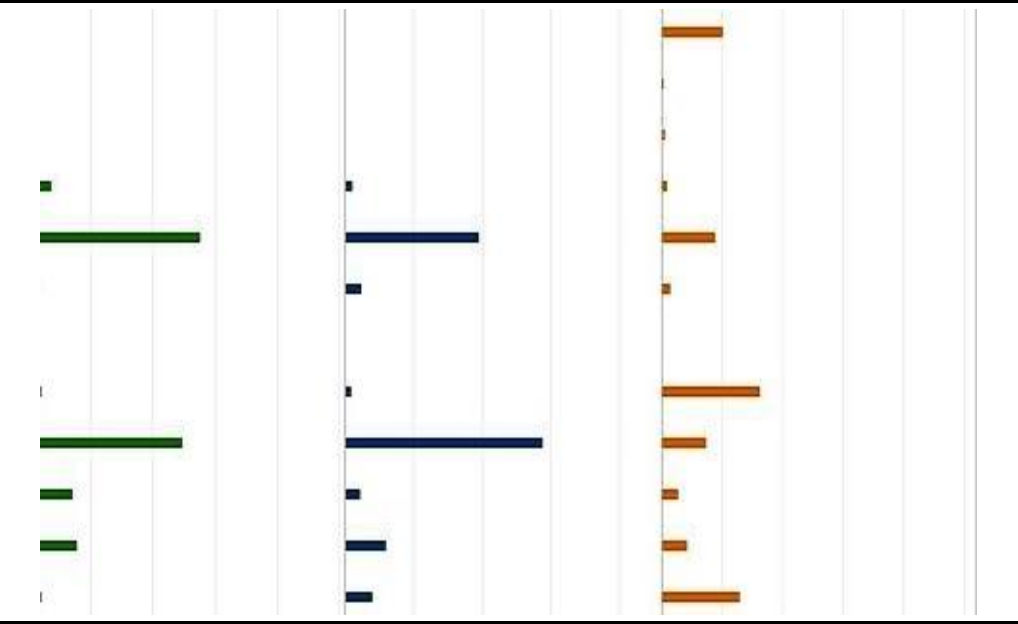 |
| | R2 | Central Asia | |
| | R3 | East Asia | |
| | R4 | Eastern Europe | |
| | R5 | Middle East & North Africa | |
| | R6 | North America | |
| | R7 | Oceania | |
| | R8 | South America | |
| | R9 | Southeast Asia | |
| | R10 | Sub-Saharan Africa | |
| | R11 | South Asia | |
| | R12 | Western Europe | |
| | | | |
| Weapon Type | WT-1 | Explosives-Bombs | 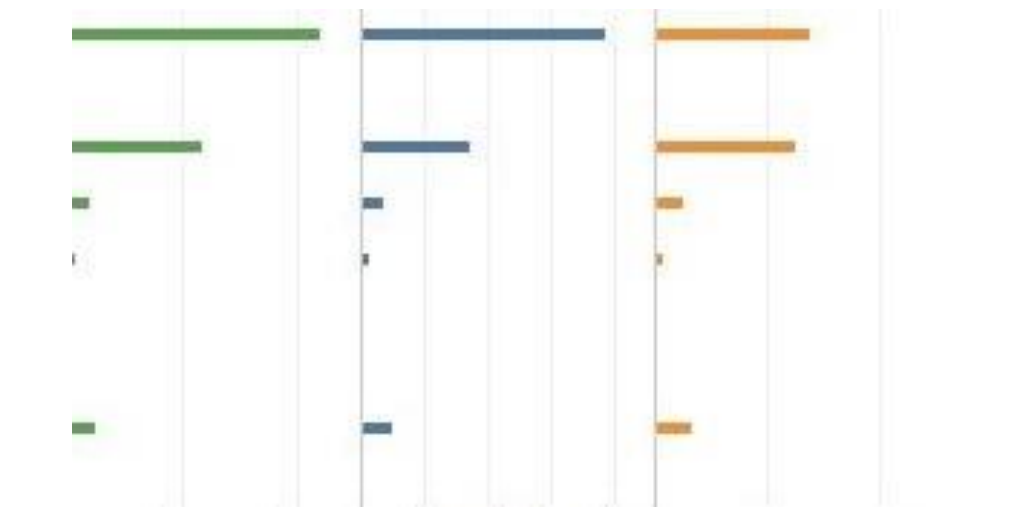 |
| | WT-2 | Fake Weapons | |
| | WT-3 | Firearms | |
| | WT-4 | Incendiary | |
| | WT-5 | Melee | |
| | WT-6 | Miscellaneous | |
| | WT-7 | Sabotage Equipment | |
| | WT-8 | Unknown | |
| | WT-9 | Vehicle | |
| Timeline | T-1 | 1970-75 | 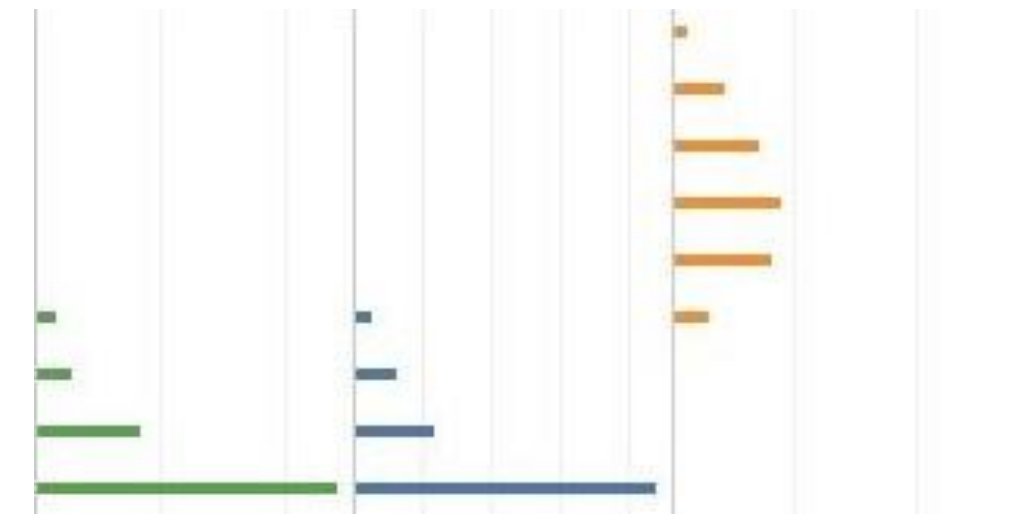 |
| | T-2 | 1976-80 | |
| | T-3 | 1981-85 | |
| | T-4 | 1986-90 | |
| | T-5 | 1991-95 | |
| | T-6 | 1996-00 | |
| | T-7 | 2001-05 | |
| | T-8 | 2006-10 | |
| | T-9 | 2011-15 | |
| Target | TG-A | Airports & Aircraft | 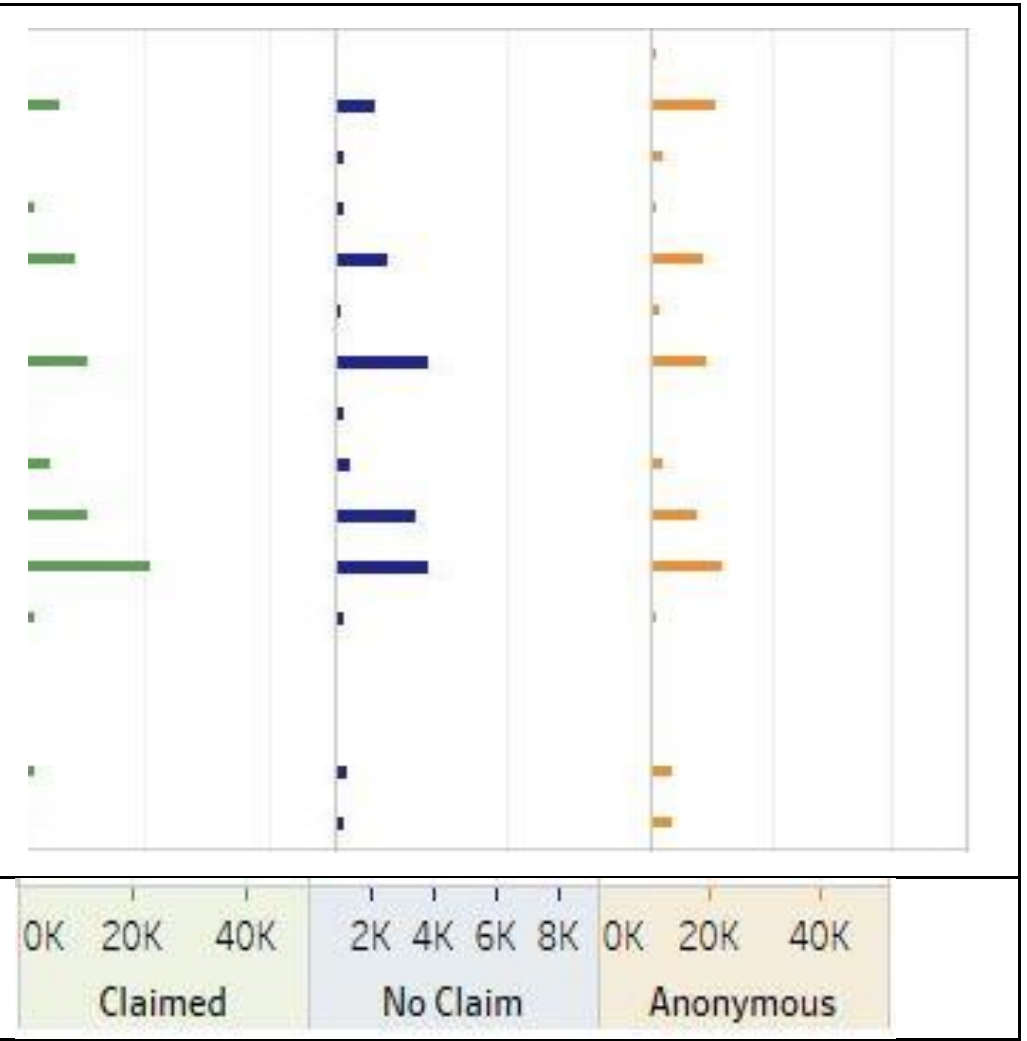 |
| | TG-B | Business | |
| | TG-C | Diplomats | |
| | TG-D | Educational Institutions | |
| | TG-E | Government Personal | |
| | TG-F | Journalists & Media | |
| | TG-G | Military | |
| | TG-H | Non-State Militia | |
| | TG-I | Others | |
| | TG-J | Police | |
| | TG-K | Private Citizens & Property | |
| | TG-L | Religious Figures /Institutions | |
| | TG-M | Telecommunication | |
| | TG-N | Tourists | |
| | TG-O | Transportation | |
| | TG-P | Utilities | |

For this work "attack responsibility" was the class, spread over three categories which are Claimed, Not-Claimed and Anonymous. Further categorization of data is also done for ease of classification. Table-1 sums the entire data categorization. It depicts the numerical distribution of data in tabular form for a better understanding of class and attributes for the terrorist attack dataset. All the attributes their code, values, total and other factors are provided in the table. Division of total attacks has been done on the basis of the class attack responsibilities which are claimed, not-claimed and anonymous.

## 4 Results and Discussion

The classifiers used for this data are: Lazy classifier IBK linear NN, Lazy classifier IBK Filtered Neighbor Search, Lazy classifier IBK, Ball Tree, Lazy classifier K-star, Decision Tree Random Forest, Multilayer Perceptron, Multiclass Classifier and Naïve Bayes. We attained results with high-end accuracies ranging from 90-95 %. Figure 5 shows the performance result of used classifiers. Tables 2 to 7 show the outputs achieved by Lazy Classifier IBK, Lazy Classifier K-star. Decision Tree Random Forest, Multilayer Perceptron, Multiclass Classifier and Naïve Bayes. This study endeavored to classify global terrorist attacks by utilizing different text mining classifiers such as Decision Tree random forest, Lazy classifier IBK linear NN, Lazy classifier IBK Filtered Neighbor Search and Lazy classifier K-star. We achieved fair accuracies in the context of classifiers performance. This analysis can be used to draw patterns for extracting information about the terror attacks. Some of the outcomes of the work are as follows:

- Out of total reported attacks from 1970 to 2015 which are 156772, 14664 are claimed, 75966 are Not-Claimed and 66142 are anonymous.

- The ratio of claiming the attacks by terrorist organizations varies from 10-15 percent. In this study, we have achieved a rate of 11%. (Although it is justified that more attacks happened from 2016 to present date and more responsibilities have been taken of those attacks so that is why the ratio is variable).
- The major stake is Asia in sustaining the attacks which are almost 45% of the total attacks.
- Government, Law Enforcement (Military and Police) bodies are the prime targets making up to 40% followed by Private Citizens and Property which is 22%.
- Losses occurred due to the terrorist attacks is highest for Moderate losses which are alone 64%.
- The types of attack such as bombing and armed assault are the most common which is 48.45% and 24.50% respectively.
- Explosives/Bombs and Firearms are the prominent weapon types deployed in these attacks with 50.47% and 33% respectively.

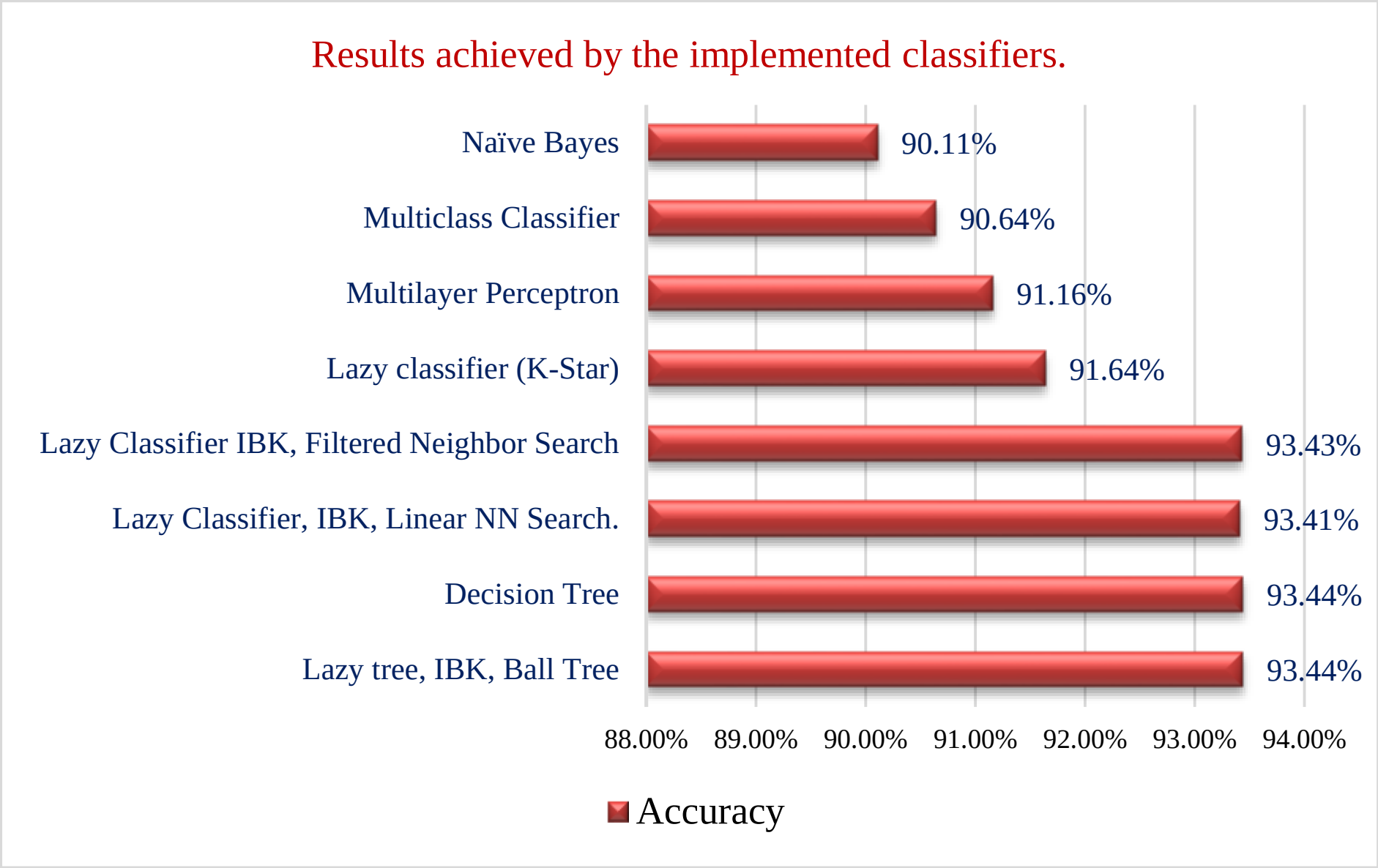

**Fig. 5.** Results achieved by the implemented classifiers.

**Table 2.** Result of Lazy Classifier IBK.

| **TP Rate** | **FP Rate** | **Precision** | **Recall** | **F-Measure** | **MCC** | **ROC Area** | **PRC Area** | **Class** |
|---|---|---|---|---|---|---|---|---|
| 0.987 | 0.987 | 0.987 | 0.987 | 0.987 | 0.987 | 0.987 | 0.987 | Claimed |
| 0.114 | 0.114 | 0.114 | 0.114 | 0.114 | 0.114 | 0.114 | 0.114 | No-Claim |
| 1.00 | 1.00 | 0.00 | 1.00 | 1.00 | 1.00 | 1.00 | 1.00 | Anonymous |

**Table 3.** Result of Lazy Classifier K-star.

| TP Rate | FP Rate | Precision | Recall | F-Measure | MCC | ROC Area | PRC Area | Class |
|---|---|---|---|---|---|---|---|---|
| 0.999 | 0.161 | 0.853 | 0.999 | 0.921 | 0.845 | 0.972 | 0.962 | Claimed |
| 0.104 | 0.00 | 0.965 | 0.104 | 0.187 | 0.303 | 0.923 | 0.572 | No-Claim |
| 1.00 | 1.00 | 0.00 | 1.00 | 1.00 | 1.00 | 1.00 | 1.00 | Anonymous |

**Table 4.** The result of Decision Tree.

| TP Rate | FP Rate | Precision | Recall | F-Measure | MCC | ROC Area | PRC Area | Class |
|---|---|---|---|---|---|---|---|---|
| 0.987 | 0.114 | 0.891 | 0.987 | 0.936 | 0.875 | 0.983 | 0.979 | Claimed |
| 0.332 | 0.005 | 0.851 | 0.332 | 0.477 | 0.509 | 0.948 | 0.669 | No-Claim |
| 1.00 | 1.00 | 0.00 | 1.00 | 1.00 | 1.00 | 1.00 | 1.00 | Anonymous |

**Table 5.** The result of Multilayer Perceptron.

| TP Rate | FP Rate | Precision | Recall | F-Measure | MCC | ROC Area | PRC Area | Class |
|---|---|---|---|---|---|---|---|---|
| **0.984** | 0.156 | 0.856 | 0.984 | 0.916 | 0.834 | 0.954 | 0.929 | Claimed |
| **0.141** | 0.008 | 0.611 | 0.141 | 0.230 | 0.268 | 0.864 | 0.389 | No-Claim |
| **1.00** | 1.00 | 0.00 | 1.00 | 1.00 | 1.00 | 1.00 | 1.00 | Anonymous |

**Table 6.** The result of Multiclass Classifier.

| TP Rate | FP Rate | Precision | Recall | F-Measure | MCC | ROC Area | PRC Area | Class |
|---|---|---|---|---|---|---|---|---|
| **0.999** | 0.180 | 0.839 | 0.999 | 0.912 | 0.828 | 0.942 | 0.910 | Claimed |
| **0.009** | 0.001 | 0.511 | 0.009 | 0.018 | 0.059 | 0.826 | 0.269 | No-Claim |
| **1.00** | 1.00 | 0.00 | 1.00 | 1.00 | 1.00 | 1.00 | 1.00 | Anonymous |

**Table 7.** The result of Naïve Bayes.

| TP Rate | FP Rate | Precision | Recall | F-Measure | MCC | ROC Area | PRC Area | Class |
|---|---|---|---|---|---|---|---|---|
| **0.986** | 0.176 | 0.841 | 0.986 | 0.908 | 0.817 | 0.939 | 0.906 | Claimed |
| **0.010** | 0.002 | 0.358 | 0.010 | 0.019 | 0.048 | 0.818 | 0.246 | No-Claim |
| **1.00** | 1.004 | 1.994 | 1.00 | 1.997 | 1.995 | 1.00 | 1.00 | Anonymous |

## 5 Conclusion and Future Work

As the analysis is user need based approach, several more patterns can be found. For future research work, there are several horizons to work upon with this analysis. First and foremost is achieving an accuracy of classifiers up to 99%. This can be done by genetic algorithms and deep neural networks along with the classifiers used in this work or a combination of the different classifier. In this study, we worked up to the tertiary classification of some of the attributes and on average up to secondary for each attribute. It has benefits and drawbacks also. Limiting the categorization of at-tributes reduces the computation complexity but for classes having a marginally low occurrence, it induces some biasing.

One of the objectives is to also increase the sub-classification layers and attributes both in order to find more useful trends. As with large data set expectation by which we mean Big Data: is to get rid of bias and variance in prediction. Secondly, using terrorist data it is possible to predict the organizations involved in the reported attacks. However, further study can be carried out to detect a terrorist group using historical data. In this lieu of work data mining, approaches can be deployed from finding the trends to counter terrorism by developing Terrorist Group Prediction Model. The parameters for the building could be web-based content, activities, social network data, phone calls, emails etc. by using the graph and pattern mining.